\documentclass[runningheads]{llncs}
\usepackage{subcaption}
\usepackage{array}
\usepackage{amssymb}
\usepackage{amsmath}
\usepackage{arydshln}
\usepackage{multicol}
\usepackage{booktabs}
\usepackage{diagbox}
\usepackage{multirow}
\usepackage{float}
\usepackage{color}
\usepackage{bbding} 
\usepackage{graphicx}
\usepackage[T1]{fontenc}
\def\doi#1{\href{https://doi.org/\detokenize{#1}}{\url{https://doi.org/\detokenize{#1}}}}
\usepackage[colorlinks,
            linkcolor=blue,       
            urlcolor=blue,
            citecolor=blue,        
            ]{hyperref}

\usepackage[table,xcdraw]{xcolor}
\usepackage{listings}
\begin{document}
%
%
\title{Self-distillation Augmented Masked Autoencoders for Histopathological Image Understanding}
\titlerunning{Self-distillation Augmented Masked Autoencoders}



\author{Yang Luo\inst{1} \and Zhineng Chen\inst{1} \and Shengtian Zhou\inst{1} \and Xieping Gao\inst{2}}

\authorrunning{Y.Luo et al.}

\institute{\textsuperscript{1}Fudan University, Shanghai, China \\
\textsuperscript{2}Hunan Provincial Key Laboratory of Intelligent Computing and Language Information Processing, Hunan Normal University \\
\email{yangluo21@m.fudan.edu.cn}\\
}

%

\maketitle              
\begin{abstract}
Self-supervised learning (SSL) has drawn increasing attention in histopathological image analysis in recent years. Compared to contrastive learning which is troubled with the false negative problem, i.e., semantically similar images are selected as negative samples, masked autoencoders (MAE) building SSL from a generative paradigm is probably a more appropriate pre-training. In this paper, we introduce MAE and verify the effect of visible patches for histopathological image understanding. Moreover, a novel SD-MAE model is proposed to enable a self-distillation augmented MAE. Besides the reconstruction loss on masked image patches, SD-MAE further imposes the self-distillation loss on visible patches to enhance the representational capacity of the encoder located shallow layer. We apply SD-MAE to histopathological image classification, cell segmentation and object detection. Experiments demonstrate that SD-MAE shows highly competitive performance  when compared with other SSL methods in these tasks.
\end{abstract}
\section{Introduction}
\label{sec:introduction}

With advancement of digital pathology, whole slide images (WSIs) scanned from glass slides have been widely applied to clinical diagnosis. Since WSIs often have ultra-high resolutions, to acquire adequate histopathological data is easy. However, to label these data can be challenging and how to utilize these resources without increasing cost has become a focal issue in histopathological image field. Recently, self-supervised learning (SSL) has received increasing research attention \cite{bert,mocov3,mae,simCLR} in natural images. SSL is an unsupervised paradigm that trains a model to learn image representations by automatically annotating data through its intrinsic structure, i.e., pretext task \cite{GeAndContra}, and fine-tunes this model in downstream tasks. Traditional SSL methods focus on different pretext tasks including \cite{doersch2015unsupervised,rotnet,zhang2016colorful}. SSL has potential to improve the performance in the downstream tasks with precisely leverage the abundance of WSIs data. Notably, it does not require manual annotation. Consequently, it has a promising application prospect in analyzing histopathological images. Typical examples include \cite{cube_rotation,nucle_size_and_num,jigsaw_puzzles}.

\begin{figure*}
     \centering
     \includegraphics[width=\textwidth]{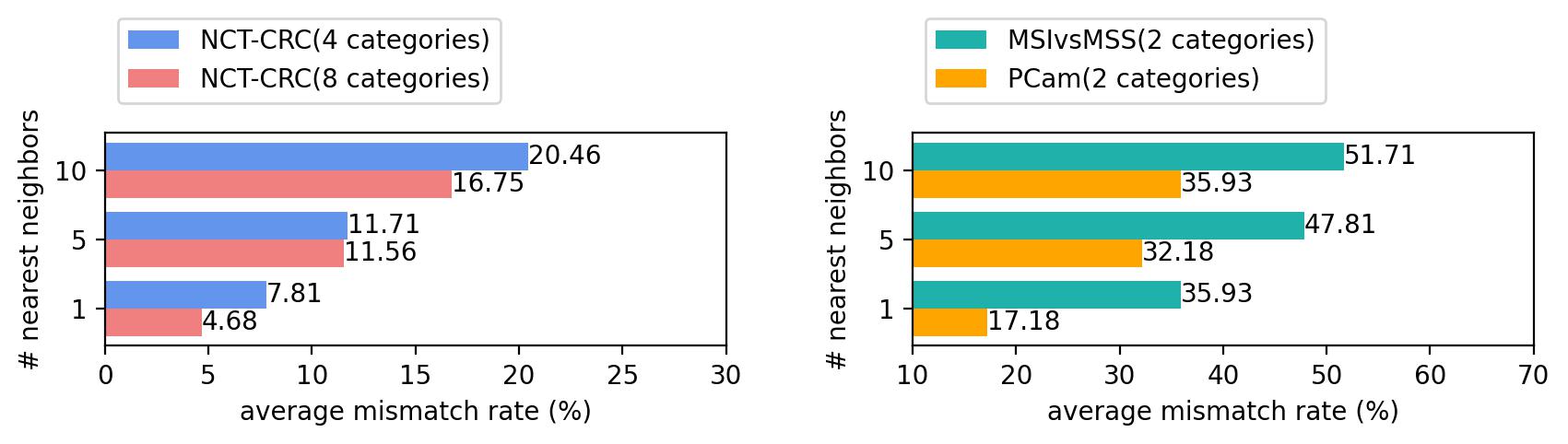}
     \caption{, \textbf{Average mismatch rate for top-k nearest neighbors} with pre-trained contrastive method (MoCo-V3) in different datasets or dataset with different categories(we keep 32 images per category).}
     \label{topk}
\end{figure*}

Recently, contrastive learning (CL) is the most prevalent paradigm of SSL. Its pre-text task is simple but effective: bringing together two views of the same image while pushing it away from other images in the feature space. Many creative works \cite{simCLR,byol,dino,mocov3}  distinguish themselves and outperform supervised methods in the natural images, and \cite{csco,wang2021transpath,ss_digital_imgs} have successfully applied it to different histopathological image analysis tasks. 

In addition, CL faces a \emph{false negative} problem, i.e., a risk of pushing away two images, despite that they belong to the same category. This results that CL highly depends on the quantity and diversity of negative samples. Some approaches alleviate this issue from diverse perspectives \cite{mocov1,mocov3,dino}. However, CL encounters two challenge within histopathological datasets. First, an image and its negative samples may have similar semantics considering that image regions are somewhat homogeneous. Second, histopathological image datasets have few classes (e.g., 2 or 4 classes \cite{PCamData,kather2019predicting}) compared to the natural image datasets (e.g., 1000 classes \cite{imagenet}). Both of the two characteristics lead to a higher likelihood of false negative in histopathological image datasets. As shown in Fig. \ref{topk}, we find that the semantic discrimination among features obtained from CL is highly sensitive to the dataset and class number. For example, in a dataset with few classes (MSIvsMSS), the average mismatch rate is obviously higher than a dataset with more classes (NCT). This indicates that CL struggles to identify negative samples in the former dataset, leading to a higher false negative risk. This phenomenon is observed in natural images \cite{scan} but is more prominent in histopathological image datasets. Overall, contrastive learning is not the optimal choice for histopathological images if the false negative problem is not carefully considered.

Recently, He \textsl{et al}. \cite{mae} proposed a new SSL paradigm termed masked autoencoders (MAE). It learns representation by a mask-and-prediction pretext task. Specifically given an image, it uses a few randomly selected visible patches to reconstruct the other masked patches. MAE relies solely on the image content for representation learning. This design can avoid the issues mentioned before. Thus it is more appropriate to the histopathological image analysis.


Motivated by the analysis above, we introduce MAE to histopathological image analysis. And we observe that MAE only requires reconstructing masked patches as similar as the raw ones, but ignores the effect of visible patches, which have the potential to improve the pre-training of the MAE encoder if applied correctly, and benefiting downstream tasks. As described in \cite{zhang2019teacher}, the position relationship between the encoded and decoded features represents exactly the shallow and deep features of the visible patches. Meanwhile, the loss function used for self distillation \cite{zhang2019teacher} can effectively enhance the shallow network. 

Motivated by it, we develop SD-MAE, a self-distillation augmented MAE. Specifically, the feature vectors of the visible patches obtained from the encoder are treated as the student whereas their counterpart from the decoder as the teacher. It can further stimulate the potential of the encoder located at the same shallow layer. Extensive experiments are conducted on downstream tasks including histopathological image classification, cell segmentation and object detection in six public benchmarks in total. The results demonstrate that SD-MAE learns a more powerful feature representation than MAE and other SSL methods.

\begin{figure}
         \centering
         \includegraphics[width=\textwidth]{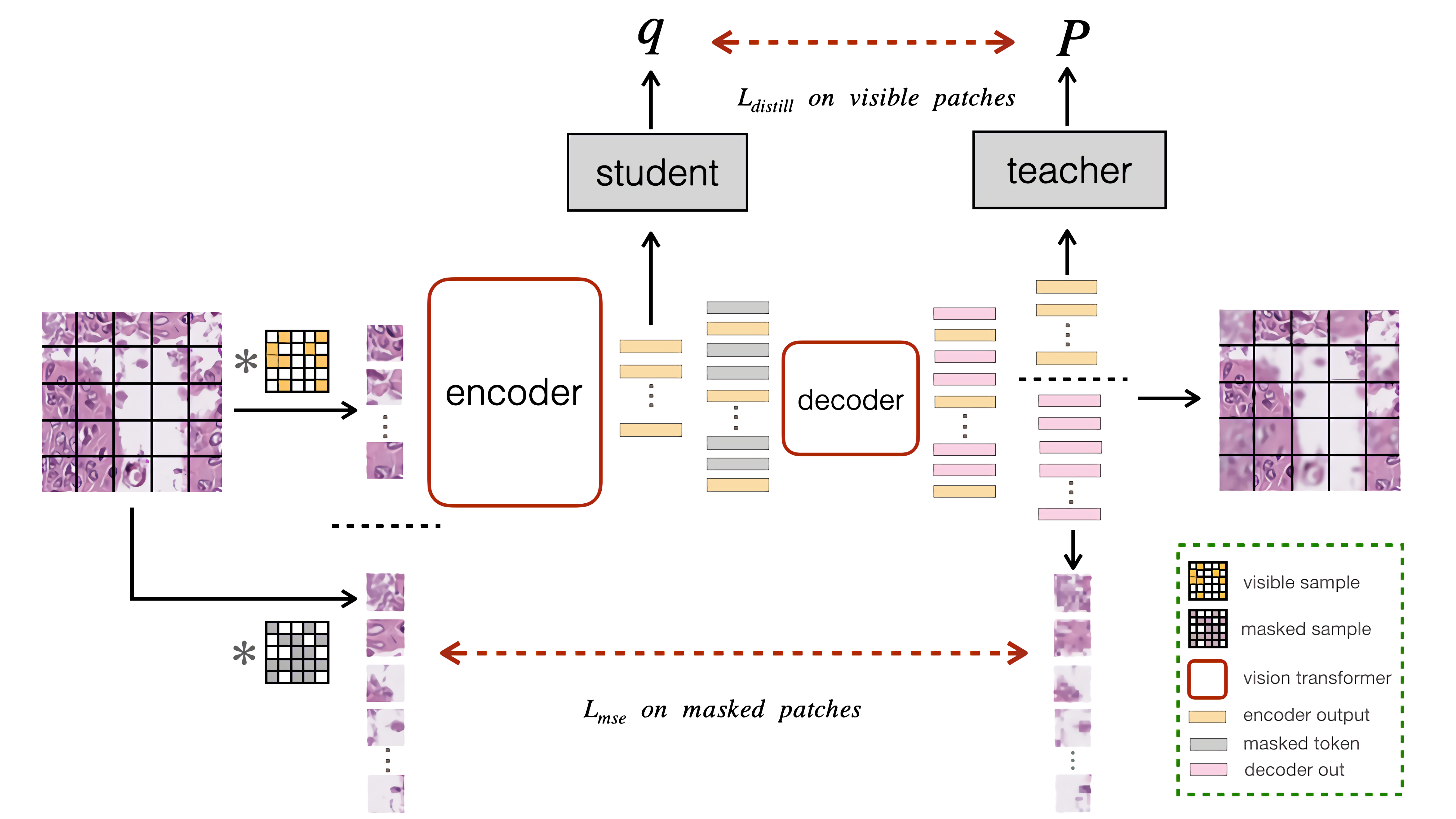}
         \caption{\textbf{Illustrative framework of the proposed SD-MAE}. Besides the raw MAE, we make use of visible patches. The decoded features are regarded as the teacher to transfer knowledge to their counterparts after encoding.}
         \label{idea}
\end{figure}

\section{Methodology}
An illustrative framework of the proposed SD-MAE is shown in Fig. \ref{idea}. It has two modules, i.e., masked image model and self distillation. The former aims to enable a generative SSL on unlabeled data by constraining the masked image patches, whereas the latter further imposes self-distillation constraints on features of visible patches to guide a more effective encoder learning.
\subsection{Preliminaries} 



MAE use the standard Vision Transformer(ViT) as the encoder. For an image  \(\mathbf{x} \in \mathbb{R}^{H \times W \times C} \), the encoder reshapes it into a sequence of flattened 2D patches \(\{x_{i}\}_{i=1}^{N}\), where \( N=HW/P^2\) and (\( P, P \)) is the resolution of each image patch. Then, patches are divided according to a masking ratio \( \mu \) into masked patches \(\{x_{m}^{i}\}_{i=1}^{\mu N}\) and visible patches \(\{x_{v}^{i}\}_{i=1}^{(1-\mu) N}\) . Visible patches are then linearly projected to visual tokens \(\mathbf{Z}_v \in \mathbb{R}^{(1-\mu)N\times D} \), where  \(D=P^{2}C\). Same as ViT, 1D position embeddings \( \mathbf{E}_{pos} \) are added to retain spatial position structure of patches. Then MAE only feeds visible tokens with the position embeddings into the encoder, and the output \(\mathbf{F}_v\) along with shared learnable mask tokens \(\mathbf{F}_m\) are fed into the decoder. Decoder output corresponding to masked patches \(\mathbf{Y}_m\) are used to compute loss. MAE uses MSE loss as the metric function \(\mathcal{L}\) and normalized pixel values of each masked patch as the prediction target. The process can be formulated as:
\begin{gather}
\mathbf{x}_m, \mathbf{x}_v =\mathrm{ RandomMask}(\mathbf{x}, \mu) \nonumber \\
\mathbf{Z}_v = \mathrm{Proj}(\mathbf{x}_v);  \\
\mathbf{F}_v  = \mathrm{Encoder}(\mathbf{Z}_v + \mathbf{E}_{pos,v}) \nonumber \\ 
\mathbf{Y} = [\mathbf{Y}_m;\mathbf{Y}_v]  = \mathrm{Decoder}([\mathbf{F}_m;\mathbf{F}_v] + \mathbf{E}_{pos})\nonumber
\end{gather}
\begin{equation}
    \mathcal{L}= \sum_{} \mathrm{MSE}(\mathrm{Norm}(\mathbf{x}_m), \mathbf{Y}_m)
\label{eqn1}
\end{equation}




\subsection{Rethinking Visible Patches}
The experiments of \cite{mae,xie2021simmim} demonstrate that the loss simply computed on all image patches (including both visible and masked ones) may have a detrimental effect. To verify the influence of including visible patches in loss calculation, we reformulate the original MSE loss into a weighted sum of two parts. We term this separation as decoupling. Specifically, the original MAE loss is only computed on masked patches, as shown in \eqref{eqn1}. Our loss function is as follows:


\begin{equation}
\begin{split}
 \mathcal{L} = \sum_{\mathbf{x} \in \mathcal{D}} (1-\alpha)* \mathrm{MSE}(\mathrm{Norm}(\mathbf{x}_m), \mathbf{Y}_m) + \alpha *\mathrm{MSE} (\mathrm{Norm}(\mathbf{x}_v), \mathbf{Y}_v)
\end{split}
\label{eqn4}
\end{equation}
where \(\alpha\) is the ratio used to control the participation of visible patches in the MSE loss. When \(\alpha\) is 0, Eq.\eqref{eqn4} is converted to Eq.\eqref{eqn1}. And when \(\alpha\) is \textit0.5, it becomes similar to experiments of \cite{mae,xie2021simmim}.  We designed a series of experiments with different \(\alpha\) to verify its effectiveness. As shown in supplementary materials, we clearly find that this decoupling (when \(\alpha\) greater than 0 and less than 0.5) is better than or comparable to the MSE loss only computed on masked patches for histopathological image classification.

While the aforementioned experiments effectively confirm that visible patches can yield beneficial effects on downstream tasks, we observe that the optimal weight ratio is not stable. Thus, we aim to explore a better way to leverage these features.

\subsection{Self-Distillation Augmented Visible Patches} 

Zhang \textsl{et al}. \cite{zhang2019teacher} introduce self distillation to transfer the knowledge of deeper layers into shallow ones within a network. This design is consistent with our objective that we expect to enhance the representation of the encoder located at the shallow layer. Motivated by it, we propose self-distillation augmented masked autoencoders (SD-MAE). Specifically, there are two kinds of latent representation vectors for visible patches in MAE, namely \( \mathbf{Z}_{v}\) after encoding and \(\mathbf{Y}_{v}\) after decoding. We treat them as shallow and deep features in the self-distillation framework \cite{zhang2019teacher}. We use a student network \( \mathcal{N}_s\) and a teacher network \(\mathcal{N}_t\) over these two vectors, resulting in two distributions \(\mathbf{q}\) and \(\mathbf{p}\), respectively. We learn to match these two distributions:
\begin{gather}
\mathcal{L}_{distill} = -\sum_{x \in \mathcal{D}}\mathbf{p(x)}\mathrm{log}(\mathbf{q(x)});\mathbf{q} = \mathcal{N}_s(\mathbf{Z}_{v}), \;\; \mathbf{p} = \mathcal{N}_t(\mathbf{Y}_{v}) 
\end{gather}
Then, the total loss is formulated as follows:
\begin{equation}
\mathcal{L} = (1-\beta ) * \mathcal{L}_{mse} + \beta * \mathcal{L}_{distill}
\end{equation}
where \(\beta \) is the empirically determined scaling factor (in our work \( \beta=0.2\)) and \(\mathcal{L}_{mse}\) is MSE loss on masked patches.





\begin{table}[htb]
\centering
\caption{\textbf{End-to-end fine-tuning evaluation results} on PCam and NCT. * are results from original papers. \textit{c} denotes the number of classes.}
\label{final}
\scalebox{0.9}{
\begin{tabular}{ccclcccccc} 
\toprule[1.2pt]
\multirow{3}{*}{Methods} & \multirow{3}{*}{Arch.} & \multirow{3}{*}{Param.} & \multirow{3}{*}{}    & \multicolumn{6}{c}{Datasets and Metrics}                                                                                                                               \\ 
\cline{5-10}
                         &                        &                         &                      & \multicolumn{3}{c}{PatchCamelyon(c=2)}                                                                    & \multicolumn{3}{c}{NCT(c=8)}                                     \\ 
\cline{5-7}\cline{9-10}
                         &                        &                         &                      & F1-score                          & Acc                          & AUC                               &  & F1-score                     & Acc                           \\ 
\hline
\hline
Supervised~              & ViT-S                  & 21                      &                      & 79.2\( \pm \) 2.9      & 79.8\( \pm \) 2.6 & 93.1\( \pm \) 0.6      &  & 89.6\( \pm \) 0.3 & 92.1\( \pm \) 0.6  \\ 
\hdashline
CS-CO \cite{csco}*                    & ResNet18               & 22                      &                      & -                                 & -                            & -                                 &  & 89.0\( \pm \) 0.3 & 91.4\( \pm \) 0.3  \\
TransPath\cite{wang2021transpath}                & ViT-S                  & 21                      & \multicolumn{1}{c}{} & 81.0\( \pm \) 1.3      & 81.2\( \pm \) 1.2 & 91.7\( \pm \) 0.4      &  & 89.9\( \pm \) 0.6 & 92.8\( \pm \) 0.8  \\
MoCo-V3\cite{mocov3}                  & ViT-S                  & 21                      &                      & 86.2\( \pm \) 2.3      & 86.3\( \pm \) 2.2 & 95.0\( \pm \) 0.6      &  & 92.6\( \pm \) 0.4 & 94.4\( \pm \) 0.4  \\
DINO \cite{dino}                     & ViT-S                  & 21                      &                      & 85.6\( \pm \) 0.6      & 85.8\( \pm \) 0.5 & 95.7\( \pm \) 0.3      &  & 91.6\( \pm \) 0.6 & 94.4\( \pm \) 0.1  \\
MAE \cite{mae}                     & ViT-S                  & 21                      &                      & 86.2\( \pm \) 1.0      & 86.7\( \pm \) 1.3 & 95.8\( \pm \) 0.3      &  & 92.4\( \pm \) 1.0 & 94.7\( \pm \) 0.5  \\
SD-MAE                   & ViT-S                  & 21                      &                      & \textbf{\textbf{87.8~$\pm$ 0.8}} & \textbf{88.2$\pm$0.5}            & \textbf{\textbf{96.2~$\pm$ 0.2}} &  & \textbf{93.5 $\pm$ 1.0}     & \textbf{95.3 $\pm$ 0.4}      \\
\bottomrule[1.2pt]
\end{tabular}
}
\end{table}

\section{Experiments And Results}
\subsection{Datasets}
To evaluate its effectiveness, we use six public datasets, PatchCamelyon (PCam) \cite{PCamData}, NCT-CRC-HE (NCT) \cite{kather2019predicting}, MSIvsMSS \cite{kather_jakob_nikolas_2019_2530835}, MoNuSeg \cite{monuseg}, Glas \cite{glas} and NuCLS \cite{glas}, first two for self-supervised learning and all for dowmstream task. For NCT, we exclude images belonging to the background in both training and validation sets following \cite{csco,wang2021transpath}. We do not divide these dataset again, and the reported results are from the validation set or the test set according to their official division.

\subsection{Experimental Setup}
We follow the same protocol in MAE \cite{mae}.  Like \cite{mocov3,xie2021simmim,mae,maskfeat,csco}, both pre-training and fine-tuning are carried out on the same dataset.  We use ViT-S \cite{dino} as the encoder and a lightweight decoder (4 transformer blocks with dimension 192 and a linear projection layer).  In pre-training, the masking ratio is 0.6. We apply L2-normalization bottleneck \cite{dino} (dimension 256 and 4096 for the bottleneck and the hidden dimension, respectively) as the projection head in self distillation. Our SD-MAE adopt 100-epoch training and 5-epoch warm-up. All experiments are carried out on 4 NVIDIA GTX 3090 GPUs. For a fair comparison, we use respective pre-training methods in\cite{dino,mocov3,wang2021transpath,mae} to train the models , and then use a unified approach \cite{mae} to fine-tune them.

\begin{table}
\centering
\caption{\textbf{Ablation studies}. First row is original MAE. Second is MAE with loss operating on all patches. Others are improvements with decoupling, different targets, self-distillation. \textit{SD} denotes self distillation loss and \textit{Feature} denotes the features after encoding of visible patches.}
\label{increment}
\begin{tabular}{cccclcccc}
\toprule[1.2pt]
\multirow{2}{*}{Method} & \multicolumn{3}{c}{Masked part}                        & \multicolumn{1}{c}{} & \multicolumn{3}{c}{Visible part}               & \multicolumn{1}{l}{\multirow{2}{*}{Acc}}  \\ 
\cline{2-4}\cline{6-8}
                        & Target                  & Loss                 & Prop. & \multicolumn{1}{c}{} & Target          & Loss                 & Prop. & \multicolumn{1}{l}{}                      \\ 
\hline
\hline
MAE     & Pixel     & MSE   & 1     & \multicolumn{1}{c}{}      & -                 & -         & -         & 86.7 \\
\hdashline
MAE     & Pixel     & MSE   & 0.5     &                           & Pixel             & MSE       & 0.5         & 85.9 \\
\hdashline
MAE     & Pixel     & MSE   & 0.8   &                           & Pixel             & MSE       & 0.2       & 87.2 \\
MAE     & Pixel     & MSE   & 0.8   &                           & Feature   & MSE       & 0.2       & 87.5 \\ 
SD-MAE  & Pixel     & MSE   & 0.8   &                           & Feature   & SD        & 0.2       & \textbf{88.2} \\
\bottomrule[1.2pt]
\end{tabular}
\end{table}

\subsection{Results and Comparisons}

\textbf{Comparisons with Previous Results.} Tab. \ref{final} reports the top-1 accuracy of different methods on the two classification datasets. As the supervised baseline, ViT-S is directly trained without pre-training. We observe that, as a classical contrastive learning method, MoCo-V3 has a high standard deviation (\( \pm \) 2.3) when the dataset has fewer classes. This is consistent with \cite{problem_contra} that contrastive learning requires more sophisticated design for histopathological datasets. In addition, Dino without negative samples has more stable performance (standard deviation \( \pm \) 0.6). MAE with image reconstruction as pretext task is more stable than MoCo-V3 (\( \pm \) 1.0 \textit{vs.} \( \pm \) 2.3) and has better performance than DINO (86.2 \textit{vs.} 85.6). Moreover, our SD-MAE improves accuracy by 1.6\% and 0.6\% on PCam and NCT respectively while performing more stable compared with MAE (\( \pm \) 0.8 \textit{vs} \( \pm \) 1.0).

\subsubsection{Alation Study.} In Table \ref{increment}, it is detailed how we improve the MAE progressively. As shown in row 2, we observe that simply using MSE loss on the visible patches leads to an accuracy drop of 0.8\% as same results as \cite{mae}. Utilizing decoupling with a low ratio (e.g., 0.2\%) boosts the accuracy by 0.5\% compared with original MAE (85.9 \textit{vs.} 86.7). And by changing the target from pixels to feature after encoding, decoupling can further improve accuracy by 0.3\%. Moreover, replacing MSE with self distillation loss improve performance compared with the former. Overall, based on these strategies, our SD-MAE archives huge improvement compared with MAE (88.2 \textit{vs.} 86.7).

\begin{table}
\centering
  \caption{\textbf{Performance of transfer learning} by pre-training and fine-tuning on different datasets.}
  \label{tab:transfer}
   
       \begin{tabular}{cclccc} 
\toprule[1.2pt]
\multirow{2}{*}{Mothod} & In-Domin  &  & \multicolumn{3}{c}{Cross-Domin}   \\ 
\cline{2-2}\cline{4-6}
 & \multicolumn{1}{l}{NCT\(\rightarrow\)MSIvsMSS} &                      & \multicolumn{1}{l}{NCT\(\rightarrow\)PCam} & \multicolumn{1}{l}{MSIvsMSS\(\rightarrow\)PCam} & \multicolumn{1}{l}{PCam\(\rightarrow\)MSIvsMSS}  \\ 
\hline
\hline
MAE    & 89.1           & \multicolumn{1}{c}{}  & 84.4              & 84.2              & 87.2 \\
SD-MAE & \textbf{90.3}  & \multicolumn{1}{c}{}  & \textbf{86.0}     & \textbf{88.9}     & \textbf{91.2} \\
\bottomrule[1.2pt]
\end{tabular}
\end{table}

\subsubsection{Transfer Learning. } To verify the robustness of our model in domain transfer, we apply methods on different pre-training and fine-tuning datasets. Specifically, MSIvsMSS and NCT both are colorectal cancer datasets, whereas PCam is lymph node sections of breast cancer. Based on the fact that WSIs from different tissues and organs commonly have different histopathological features (e.g., cell morphology, mode of cell composition), we consider MSIvsMSS and NCT to be within the same domain, while both of them are cross-domain with PCam. In Tab. \ref{tab:transfer}, our SD-MAE gets 1.2\% improvement over the MAE (90.3 \textit{vs.} 89.1) in the in-domin transferring. And SD-MAE archives obviously better transfer performance than MAE in the cross-domin transferring, e.g., with improvements ranging from 4\% to 5.7\%. 

\begin{figure}
\centering
\includegraphics[width=1\columnwidth]{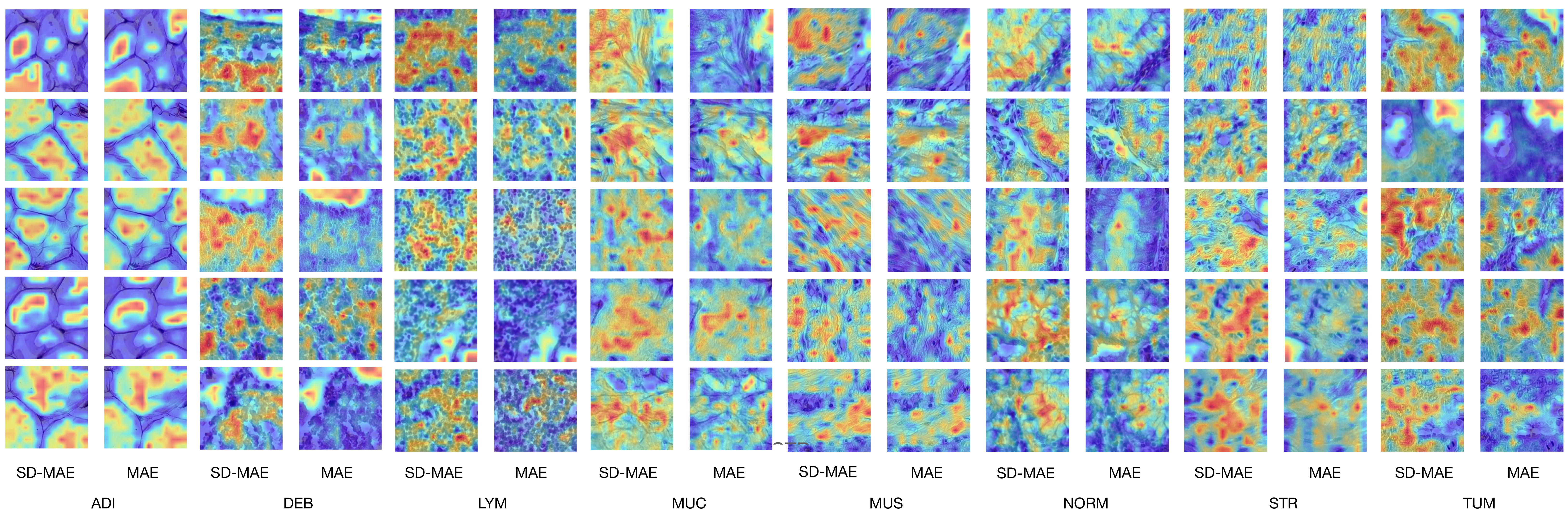}
\caption{\textbf{Attention map} of the last layer of the encoder after pre-training over the validation dataset of NCT. Words represent different tissues or diseases.} 
\label{figatten}
\end{figure}

We analyze and explain it from perspectives of feature diversity, we classify datasets into (\(\mathit{high}\) and \(\mathit{low}\)) based on feature diversity (more detailed of partition are shown in appendix). We observe that SD-MAE performs better than MAE when it transfers from \(\mathit{high}\) to \(\mathit{low}\) or from \(\mathit{low}\) to \(\mathit{low}\). For the former, we attribute it to its wide attention distribution which can help SD-MAE focus on more features. For the latter, SD-MAE has higher activation values for the same feature. This allows SD-MAE can focus on key features even pre-training with datasets of low diversity. Visualizations of attention maps are shown in Fig. \ref{figatten}



\subsubsection{Semantic segmentation.}
We perform experiments of semantic segmentation on MoNuSeg \cite{monuseg} and Glas \cite{glas}. We use ViT as backbone and UperNet \cite{upernet} as the head of decoder following the code in \cite{image_beit}. As shown in Tab. \ref{tab:segmentation}, SD-MAE consistently reports better results compared with MAE. 
\begin{table}
\centering

  \caption{\textbf{Results of semantic segementation} on MoNuSeg and Glas. aAcc is accuracy over all pixels.}
  \label{tab:segmentation}
    \centering
\begin{tabular}{clccccccccccc} 
\toprule
\multirow{2}{*}{Mothod}    & \multirow{2}{*}{} & \multicolumn{2}{c}{PCam-MoNuSeg} &  & \multicolumn{2}{c}{\;\quad PCam-Glas \quad\quad} &  & \multicolumn{2}{c}{NCT-MoNuSeg} &  & \multicolumn{2}{c}{\;\quad NCT-Glas \quad\quad}  \\ 
\cline{3-4}\cline{6-7}\cline{9-10}\cline{12-13}
        & & aAcc & mIoU & & aAcc & mIoU & & aAcc & mIoU & & aAcc & mIoU \\ 
\hline\hline
MAE     & & 92.2 & 79.3 &  & 90.4 & 82.5 &  & 91.2 & 78.0 &  & 90.1 & 82.0 \\
SD-MAE  & & \textbf{92.3} & \textbf{79.5} &  & \textbf{90.6} & \textbf{82.8} &  & \textbf{92.0} & \textbf{79.1} &  & \textbf{90.6} & \textbf{82.9} \\
\bottomrule
\end{tabular}
\end{table}

\subsubsection{Object detection.} We adopt Mask R-CNN \cite{vitdet} with ViT-S pre-trained from SD-MAE as the backbone. We report box mAP calculated at IoU threshold 0.5 on NuCLS \cite{nucls}. Compared to MAE, our methods perform better under the same configuration. When pre-trained on PCam or NCT, SD-MAE is 0.7 (19.7 \textit{vs.} 19.0)  and 1.6 (20.5 \textit{vs.} 18.9) points higher than MAE, respectively.

\section{Conclusion}
We observe that contrastive learning exhibits higher false negative risk in histo-pathological image datasets. Alternatively, we introduce MAE to histopathological image analysis and propose SD-MAE, a self-distillation augmented MAE. It transfers knowledge from the decoder to the encoder by making use of visible patches to enhance the representational capacity of the encoder located shallow layer. Extensive experiments on six public benchmarks covering different downstream tasks basically validate our proposal. Compared to MAE for histopathology images, SD-MAE guides more effective visual pre-training. In the future, we plan to further explore SD-MAE on other types of medical images.






\newpage
\bibliographystyle{IEEEtran}      
\bibliography{ebib}

\appendix

\section{Supplementary Materials}

\begin{table}
\centering
\caption{\textbf{Experiments verifying the effectiveness of decoupling}.  Detailed configurations are provided in Tab.\ref{conditions}. The top-1 accuracy of different experiments on the two datasets are reported. \(\alpha\) denotes decoupling with different ratio. }
\label{decoupling}
\begin{tabular}{cccccclccccc} 
\toprule
\multirow{2}{*}{ID} & \multicolumn{5}{c}{PCam}                                                                 & \multicolumn{1}{c}{} & \multicolumn{5}{c}{NCT}                                                                                                             \\ 
\cline{2-6}\cline{8-12}
                    & \(\alpha\)=0    &\(\alpha\)= 0.1                              & \(\alpha\)=0.2                              & \(\alpha\)=0.3  &\(\alpha\)= 0.5  &                      & \(\alpha\)=0                     & \(\alpha\)=0.1                              & \(\alpha\)=0.2  & \(\alpha\)=0.3                               &\(\alpha\)= 0.5                               \\ 
\hline
\hline
S1                  & 85.4 & \textbf{87.6} & 86.8                             & 85.4 & 85.8 &                      & 94.1                  & \textbf{94.5} & 94.1 & 94.2                              & 93.8                              \\
S2                  & 87.0 & \textbf{87.8} & 86.6                             & 86.7 & 84.1 &                      & \textbf{94.5} & 94.3                             & 94.3 & 94.4                              & 94.2                              \\
S3                  & 87.3 & 86.8                             & \textbf{88.0} & 87.7 & 86.6 &                      & 94.7                  & 93.9                             & 93.8 & \textbf{94.7} & 94.2                              \\
S4                  & 87.1 & 86.3                             & \textbf{87.2} & 86.2 & 85.9 &                      & 94.7                  & 94.6                             & 94.1 & 94.4                              & \textbf{95.3}  \\
\bottomrule
\end{tabular}
\end{table}
\begin{table}
\centering
 \caption{\textbf{Conditions} applied to prove the effectiveness of the decoupling, considering learning rate and weight decay are two key factors of training.}
    \label{conditions}
    \begin{tabular}{cccccc} 
    \toprule[1.2pt]
    \multirow{2}{*}{ID} & \multirow{2}{*}{Arch.}    & \multirow{2}{*}{Param.} & Pre-training  & \multicolumn{2}{c}{Fine-tuning}  \\ 
    \cline{4-6}
                        &                           &                         & Learning rate & Learning rate & Weigh-decay      \\ 
    \hline
    \hline
    S1                  & ViT-S                     & 21                      & 1e-4          & 5e-4          & 5e-3             \\
    S2                  & \multicolumn{1}{l}{ViT-S} & 21                      & 1e-4          & 5e-4          & 5e-2             \\
    S3                  & \multicolumn{1}{l}{ViT-S} & 21                      & 1.5e-4        & 5e-4          & 5e-2             \\
    S4                  & \multicolumn{1}{l}{ViT-S} & 21                      & 1.5e-4        & 1e-3          & 5e-2             \\
    \bottomrule[1.2pt]
    \end{tabular}
\end{table}




\begin{table}
            \centering
            \caption{\textbf{Feature diversity of dataset.} We classify the datasets into two distinct categories (\(\mathit{high}\) and \(\mathit{low}\)) based on the number of categories in tissue and disease.}
            \label{dataset_charact}
            \begin{tabular}{cllccc} 
            \toprule[1.2pt]
            \multirow{2}{*}{Dataset} &  &  & \multicolumn{3}{c}{Dataset characteristics}                                                                                                                                                                                                                                            \\ 
            \cline{4-6}
                                     &  &  & \# classes & Tissue/Disease classes                                                                                                                                                                                                                        & Feature diversity  \\ 
            \hline
            \hline
            PCam                     &  &  & 2                 & breast cancer                                                                                                                                                                                                                                 & low                \\
            NCT                  &  &  & 8                 & \begin{tabular}[c]{@{}c@{}}adipose(ADI), debris(DEB),\\ lymphocytes(LYM),  mucus(MUC), \\ smooth muscle(MUS), \\normal colon mucosa(NORM),  \\cancer-associated stroma(STR),\\ colorectal adenocarcinoma epithelium(TUM)\end{tabular}                                                                    & high               \\
            MSIvsMSS                 &  &  & 2                 & colorectal cancer, gastric cancer                                                                                                                                                                                                             & low                \\
            MoNuSeg                  &  &  & 7                 & \begin{tabular}[c]{@{}c@{}}breast invasive carcinoma,\\ kidney renal clear cell carcinoma, \\lung squamous cell carcinoma,\\ prostate adenocarcinoma, \\bladder urothelial carcinoma, \\ colon adenocarcinoma, \\stomach adenocarcinoma\end{tabular} & high               \\
            Glas                     &  &  & 1                 & colorectal adenocarcinoma                                                                                                                                                                                                                     & low                \\
            NuCLS                    &  &  & 1                 & breast cancer                                                                                                                                                                                                                                 & low                \\
            \bottomrule[1.2pt]
            \end{tabular}
\end{table}
\begin{table}
            \centering
\caption{\textbf{Comparison between SD-MAE and MAE} for transferring learning in datasets with different feature diversity. SD-MAE performs better than MAE when it transfers from \(\mathit{high}\) to \(\mathit{low}\) or from \(\mathit{low}\) to \(\mathit{low}\).}
\label{transfer_res}
\begin{tabular}{cccc} 
\hline
Pre-train data & Fine-tune data & Variation in feature diversity & Result  \\ 
\hline
NCT        & PCam           & high\(\to\)low                     & +1.6    \\
NCT        & MSIvsMSS       & high\(\to\)low                     & +1.2    \\
NCT        & Glas           & high\(\to\)low                     & +0.9    \\
NCT        & NuCLS          & high\(\to\)low                     & +1.6    \\
NCT        & MoNuSeg        & high\(\to\)high                     & +1.1    \\
PCam           & MSIvsMSS       & low\(\to\)low                      & +4.0    \\
MSIvsMSS       & PCam           & low\(\to\)low                      & +4.7    \\
PCam           & NuCLS          & low\(\to\)low                      & +0.7    \\
PCam           & Glas           & low\(\to\)low                      & +0.3    \\
PCam           & NCT        & low\(\to\)high                     & -0.3    \\
MSIvsMSS       & NCT        & low\(\to\)high                     & -0.2    \\
PCam           & MoNuSeg        & low\(\to\)high                     & +0.2    \\
\hline
\end{tabular}
\end{table}

\end{document}